\pdfoutput=1

\documentclass[11pt]{article} 
\usepackage[final]{acl}
\usepackage{times}
\usepackage{latexsym}

\usepackage[T1]{fontenc}

\usepackage[utf8]{inputenc}

\usepackage{microtype}

\usepackage{inconsolata}

\usepackage{graphicx}

\usepackage{multicol}

\usepackage{multirow}
\usepackage{microtype}
\usepackage{hyperref}
\usepackage{url}
\usepackage{booktabs}
\usepackage{array} 
\usepackage{makecell}
\usepackage{listings}
\usepackage{xspace}
\usepackage[draft]{minted}
\usepackage{subcaption}
\usepackage{graphicx}
\usepackage{amsmath}
\usepackage{tcolorbox}
\definecolor{colSc}{HTML}{FF968D}
\definecolor{colCg}{HTML}{8BF1DF}
\definecolor{colIn}{HTML}{F1CF95}

\newcommand{\scratch}{\colorbox{colSc}{\textit{Scratch}}\xspace}
\newcommand{\tweak}{\colorbox{colSc}{\textit{Tweak}}\xspace}

\newcommand{\Turtle}{TurtleBench\xspace}
\newcommand{\gpt}{GPT-4o\xspace}
\newcommand{\llava}{Llava-1.5-13B}
\newcommand{\gemini}{Gemini 1.5 Flash\xspace}
\newcommand{\rabbit}{Rabbit\xspace}
\title{\includegraphics[width=0.8cm]{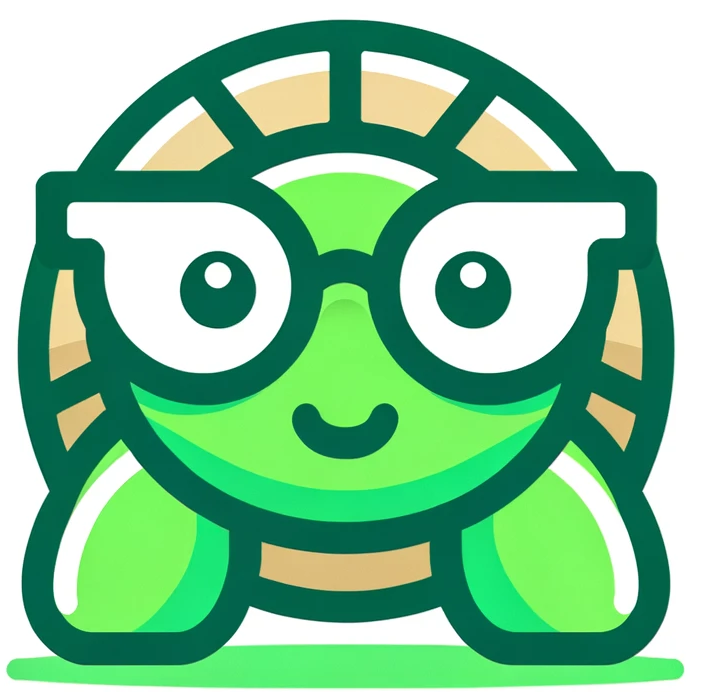} \Turtle: A Visual Programming Benchmark in Turtle Geometry}

\author{Sina Rismanchian, Yasaman Razeghi, Sameer Singh, Shayan Doroudi
\\
University of California, Irvine\\
\texttt{\{srismanc,yrazeghi,sameer,doroudis\}@uci.edu} \\
}

\begin{document}

\maketitle

\begin{abstract}
Humans have the ability to reason about geometric patterns in images and scenes from a young age. 
However, developing large multimodal models (LMMs) capable of similar reasoning remains a challenge, highlighting the need for robust evaluation methods to assess these capabilities.
We introduce \Turtle, a benchmark designed to evaluate LMMs' capacity to interpret geometric patterns---given visual examples, textual instructions, or both---and generate precise code outputs. 
Inspired by turtle geometry, a notion used to teach children foundational coding and geometric concepts, TurtleBench features tasks with patterned shapes that have underlying algorithmic logic. 
Our evaluation reveals that leading LMMs struggle significantly with these tasks, with GPT-4o achieving only 19\% accuracy on the simplest tasks and few-shot prompting only marginally improves their performance ($<2\%$).
\Turtle highlights the gap between human and AI performance in intuitive and visual geometrical understanding, setting the stage for future research in this area.
\Turtle stands as one of the few benchmarks to evaluate the integration of visual understanding and code generation capabilities in LMMs, setting the stage for future research.
Code and Dataset for this paper is provided here: \href{https://github.com/sinaris76/TurtleBench}{https://github.com/sinaris76/TurtleBench}

\end{abstract}

\section{Introduction}
Large Multimodal Models (LMMs) have the potential to handle tasks that combine visual, linguistic, and reasoning abilities, previously achievable only by humans.
Indeed, LMMs such as GPT4-V \citep{yang2023dawn} and Gemini 1.5 flash \citep{team2023gemini,fu2023challenger} have been shown to be state-of-the-art models in solving multi-modal tasks such as visual question answering \citep{goyal2017making,liu2024visual}, coding for visual multimodal questions \citep{li2024mmcode}, visual mathematical questions \citep{lu2023mathvista}, chart question answering \citep{masry-etal-2022-chartqa}, 
etc. 
Despite these successes, there remains the question of how LMMs perform in tasks that intertwine visual reasoning and programming knowledge. 
That is, given an image of a geometric pattern (and/or a verbal description of the pattern) can LMMs generate code that would be able to procedurally generate that pattern? 
Indeed, 
\citet{bubeck2023sparks} showed that the large language model (LLM) with no visual training data was able to create a unicorn in TikZ---the LaTeX-based graphics drawing library. 
This feat amazed many and provoked many discussions on the intelligence of large language models---but how general is this ability?

\begin{figure*}
    \centering      
    \includegraphics[width=1\linewidth]{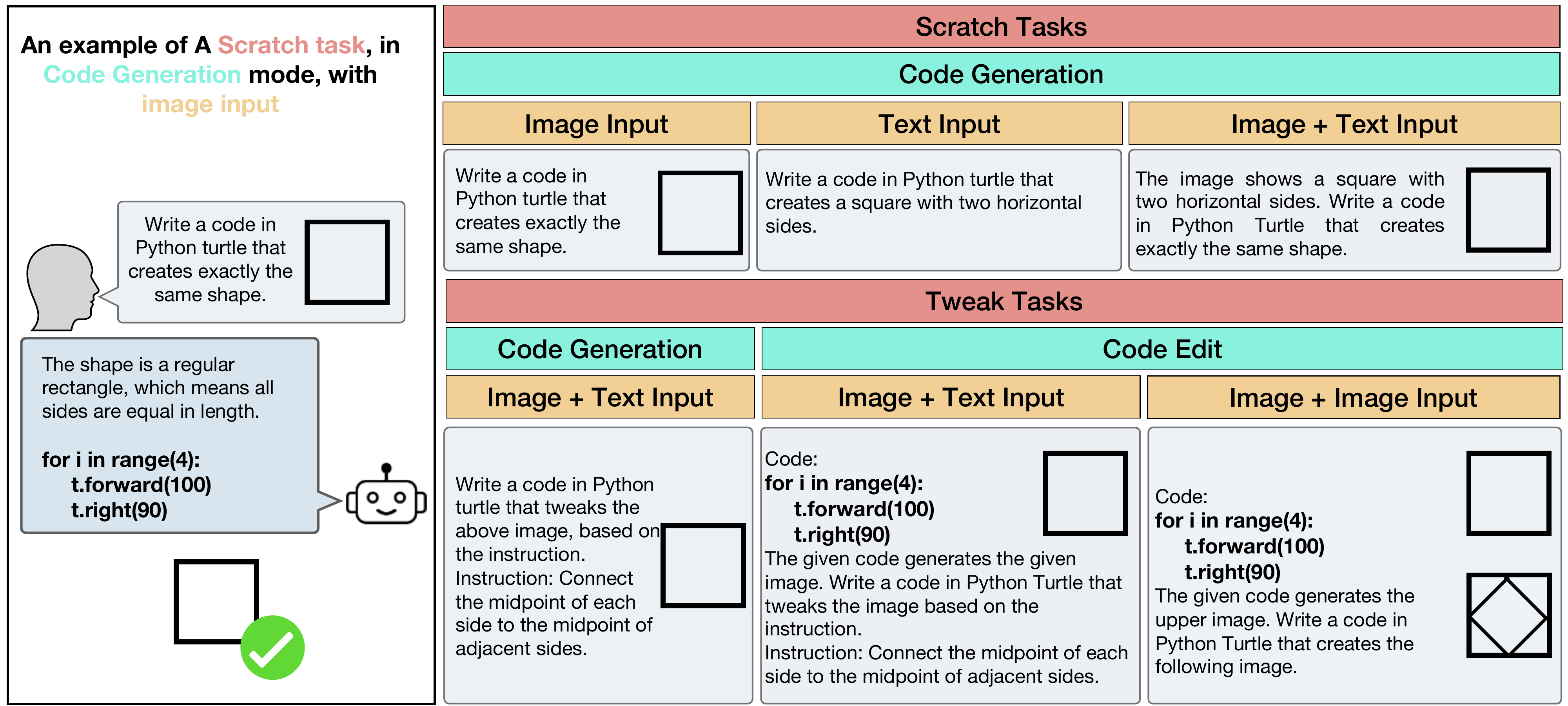}
    \caption{An illustration of existing types and modes in \Turtle. }
    \label{fig:intro}
\end{figure*}

In this work, we introduce \Turtle, a set of manually crafted image/text to code tasks in turtle geometry \citep{papert1972making,abelson1986turtle} to evaluate the abilities of these models to combine visual pattern recognition, mathematical reasoning, Python programming, and abstract geometrical reasoning.
To ensure the visual inputs and the programming language remain straightforward, \Turtle harnesses turtle geometry, a concept widely recognized for its effectiveness in introducing programming concepts to children within the K-12 education system.
In turtle geometry, a turtle acts as a programmable object that navigates the screen, drawing as it goes and turning at specified angles, to create simple visual patterns. 
The primary objective within this framework is to generate code capable of producing simple visual inputs.
These visual inputs consist of basic geometric shapes, and the programming syntax required is intentionally limited and straightforward.
An example of such a task is presented in the left side of Figure \ref{fig:intro}. 
As illustrated, the input image is the shape of a simple square and the corresponding code only uses two simple turtle functions (\texttt{forward} and \texttt{right}) along with a simple for loop.
This simplicity makes \Turtle an effective benchmark for evaluating the capabilities of Large Multimodal Models (LMMs).
To reflect different real-world use cases of an LMM in the domain of Turtle and also cover the broad range of underlying reasoning abilities, 
\Turtle{} includes 260 tasks with a variety of types and modalities.
The different types and input/output modalities are presented in Figure~\ref{fig:intro}.
We define two types of tasks \colorbox{colSc}{scratch} and \colorbox{colSc}{tweak} in \Turtle.
\colorbox{colSc}{Scratch} tasks challenge models to \colorbox{colCg}{\emph{generate}} Python code for a specified shape using the Turtle library based on inputs that could be an image of the shape, a textual description, or both. 
This subset evaluates model's proficiency in recognizing patterns within the shape and accurately translating these into executable code.
Conversely, \colorbox{colSc}{tweak} tasks are designed to probe deeper into a model's understanding, examining its ability to comprehend the implications of described modifications to shapes—such as connecting midpoints (example in Figure \ref{fig:intro})—and their representation in the image.
Here, models are provided with a base shape and are \emph{instructed} to create the desired alteration in shape. 
Instructions for these modifications may be provided \colorbox{colIn}{visually} or \colorbox{colIn}{textually}. 
To simplify, in a subset of the \colorbox{colSc}{tweak} tasks, we adopt a \colorbox{colCg}{\emph{code editing}} approach, supplying the original shape's code and directing the model to \colorbox{colCg}{\emph{edit}} this code to generate the target shape.

We conduct an evaluation of leading LMMs on \Turtle code generation and code editing tasks,  utilizing zero-shot and visual chain-of-thought approach \citep{singh2023assessing}  across text-only, image-only, and mixed (text and image) input modalities.
Our findings reveal that these models generally perform poorly across all setups and  variety of tasks and modalities. 
Both GPT-4o and Gemini 1.5 Flash struggle with \Turtle tasks, failing to solve more than 75\% of them.
Our results show that performance improves when tasks are presented as text rather than images, suggesting that integrating visual and linguistic information, especially for visual pattern recognition, requires further refinement. 
When tested with a custom library mimicking Python Turtle but using different command names, models showed a significant performance drop, revealing difficulties in generalizing visual reasoning to unfamiliar syntax. 
Even when allowed to choose their own programming language, models consistently failed to generate correct code, indicating broader challenges in translating visual instructions into functional programming outputs.
These findings show that our benchmark poses a significant challenge for LMMs, offering key insights into their limitations. Our evaluation highlights gaps in integrating visual reasoning with programming and raises important questions for future research to address.

\section{\includegraphics[width=0.7cm]{figs/turtle.png} Overview of \Turtle{}}

In turtle geometry \citep{abelson1986turtle}, a turtle is a programmable object on the screen that leaves a trace while moving. 
As illustrated in Figure \ref{fig:intro} left side, we see an example of how creating a simple geometric shape, like a square, involves the turtle moving forward and executing turns four times.
It is a powerful, intuitive tool that enables novice learners to start learning programming by creating aesthetically beautiful artifacts.
Although Turtle programming nowadays is used more as a tool to foster computational thinking, it can be used to teach geometry and mathematical reasoning \cite{marji2014learn,clements2013children} as it enables learners to explore and learn geometrical relationships between shapes. 
The intuitive nature and learnability of turtle geometry, along with its ability to generate patterns of diverse complexities, make it a compelling concept upon which to base a benchmark for LMMs.
In the following, we describe our benchmark in detail.

\subsection{\Turtle{} Task Types}
\Turtle{} is a set of 260 tasks that are designed to evaluate LMMs' performance on vision and language algorithmic reasoning tasks.
To ensure the novelty of the tasks and their quality in incorporating authentic geometric shapes and concepts, we craft \Turtle manually. 
All the tasks in \Turtle are accurately solvable based on the provided information for each, which means that there are no ambiguities or arbitrary parameters leading to inaccuracies in the tasks for humans as well as the models. To remove possible ambiguities in the tasks, two independent annotators worked with us to identify and resolve any unclear instructions.
Each task consists of a black-and-white image illustrating a set of abstract geometric shapes as an \emph{input}. 
An example of this task is presented in Figure \ref{fig:intro}.
\Turtle{} is made up of two different types of tasks, these types reflect the methodologies used in turtle geometry to introduce programming to children:
\colorbox{colSc}{Scratch} tasks that are intended to show how well a model understands a pattern and translates its understanding to an executable code.
In the general case of this type of task, an image is provided, and the requested output is code in Python Turtle that creates the shapes in the image.
In all scratch tasks, the model is asked to \emph{generate} the code in Python Turtle for the desired input shape.
\Turtle includes a total of $130$ scratch tasks and $130$ tweak tasks resulting in $260$ tasks overall.
An example of these tasks is provided in Figure \ref{fig:intro} top rows. 
To distinguish between the models' visual comprehension and their textual understanding, a subset of these tasks includes a text description of the image input in addition to the visual representation.
This setup facilitates the evaluation of how models respond differently to visual and textual inputs, providing a clearer understanding of their capabilities.
\colorbox{colSc}{Tweak} tasks that are intended to measure how well a model uses their understanding of a visual pattern, combined with an instruction to make minimal alterations. 
Each tweak task presents a model with an image and an instruction; the expected output is Python Turtle code that modifies the shape in the input image according to the given instruction.
These tasks are particularly insightful for determining whether a model is merely recalling memorized code for an image, or if it has developed a deeper, more human-like comprehension of the patterns depicted in the images. 
For instance, a model might be capable of generating code for a certain shape based on training data, but the real challenge lies in its ability to adapt that shape in response to various instructed changes. 
An example of these tasks is provided in Figure \ref{fig:intro} bottom row. 
Here, The model is given an input image of a rectangle, with an instruction to \emph{connect the midpoint of each side to the midpoint of adjacent sides}.
As illustrated in Figure \ref{fig:intro}, we also introduce a code editing version of the tweak task. 
In this version, we supply the code corresponding to the input image and then instruct the models to make specific modifications to this code, aiming to achieve a change in the image as per the provided instructions.
Detailed information about types of tweaks and their examples is provided in Appendix~\ref{app:tweaks}.

\begin{figure*}
    \centering
    \includegraphics[width=\linewidth]{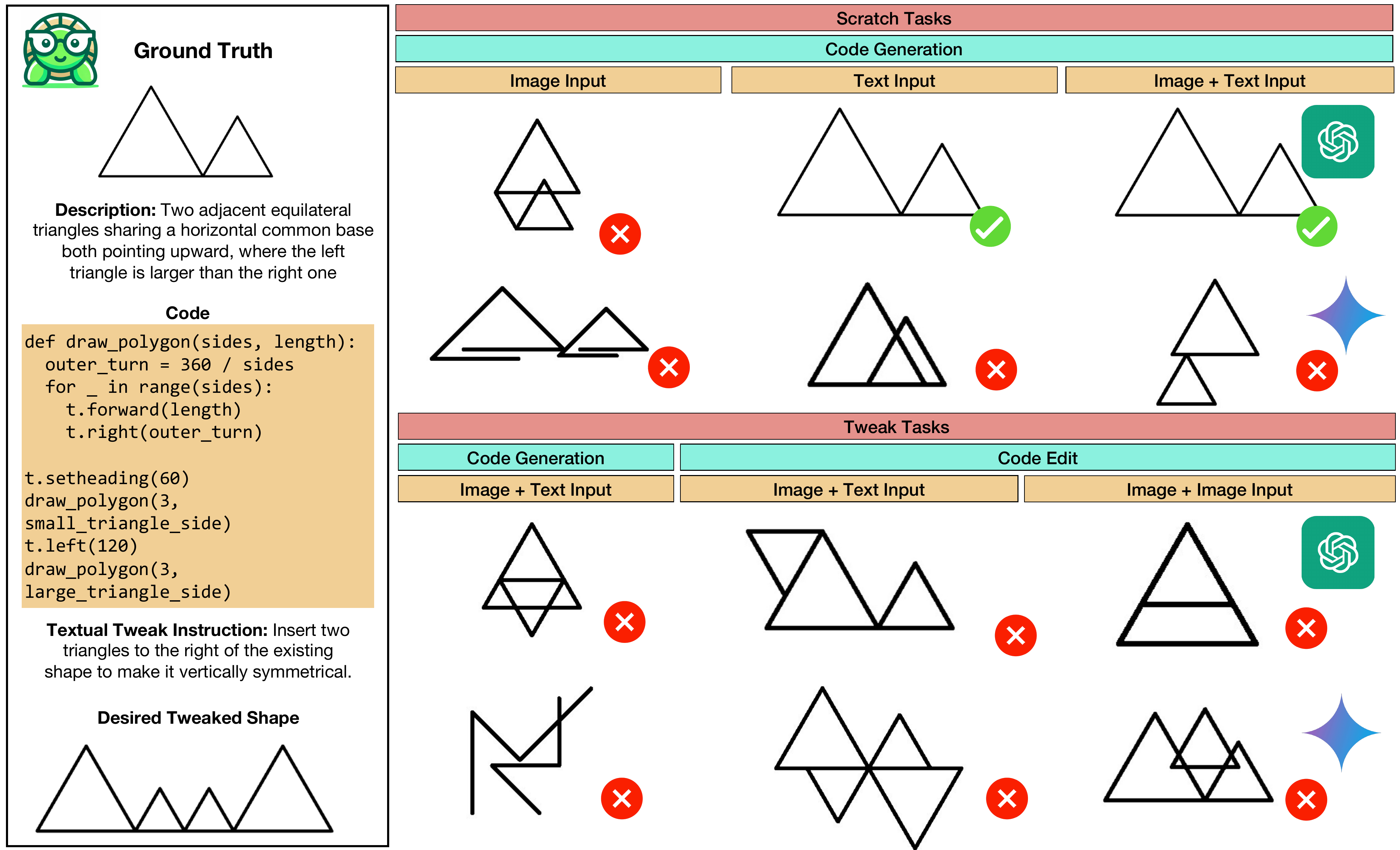}
    \caption{An illustration of different modes of a single task in \Turtle along with the images generated by code from the outputs of \gpt and \gemini. 
    More examples are provided in Appendix Figure \ref{fig:example2}}
    \label{fig:eval_examples}
\end{figure*}

\subsection{Automatic Evaluation of Code Output}
Evaluation of the output code by an AI model is performed automatically. 
First, the output of the AI model is processed to extract the code piece of output. 
Then, this piece of code is run in a sandbox, and the shape produced by the code is stored.
An illustration of this pipeline is provided in Figure \ref{fig:eval_pipeline}.
Finally, using the \textit{OpenCV} module in Python, the binary versions of the correct shape and the produced shape are compared using an adjusted measure of bitwise similarity where we first use the bounding box technique with \textit{OpenCV} to find the exact location of the shape and then calculate similarity with the formula:

\[
\frac{|B_a \cap B_m|}{|B_a \cup B_m|}
\]

where \( B_a \) and \( B_m \) represent black pixels in the input and LMM output, respectively. This metric measures the ratio of co-occurring black pixels to the total black pixels
Here, we utilize a heuristic approach in labeling the correctness of the model's output.
If the bitwise similarity between output and ground truth is higher than 95\% the models' output is labeled as correct and incorrect otherwise. 
To make sure that our heuristic in labeling the correctness of generated shapes is reliable, we manually annotated 2000 pairs of input and output images and we found that only three instances of pairs were labeled incorrectly (two of them false negative and the other false positive.), leading to an error rate of $0.15\%$ which shows the high level of reliability in the heuristic we used.

\section{Evaluation Setup}
\subsection{Models} In the following section, we evaluate \Turtle using two SOTA LMMs, \gpt and \gemini and also an open sourced model, namely \llava{} \citep{liu2023improved} employing greedy decoding in our evaluations. 
We evaluated two other open models, namely Qwen-VL-Max \citep{bai2023qwen} and CogVLM \citep{wang2023cogvlm} on a subset of tasks in \Turtle. 
However, CogVLM and Qwen are not successful in producing a syntactically correct Python Turtle piece of code even for the simplest tasks, therefore we limited our benchmark evaluation to models mentioned above. \\
\subsection{Prompting} 
We use two types of prompting in our experiments, 1) basic, where we simply prompt the the model (c.f. Appendix~\ref{app:prompt}) to do our tasks.
, and 2) Chain-of-Thought (CoT) prompting \citep{wei2022chain}, which has shown to be an effective prompting technique in eliciting reasoning in these models.
Specifically, we use a more detailed version of CoT prompting that is tailored to LMMs, namely v-CoT, recently proposed by \cite{singh2023assessing}.
The v-CoT approach is inspired by m-CoT \citep{zhang2023multimodal}, which shows higher performance compared to it. 
This prompting has been shown to improve LMMs' performance on visual tasks that involved reasoning, such as ARC \citep{chollet2019measure}. 
This prompt, instructs the model to first extract all the relevant information in the image needed for answering the problem and then to reason step by step based on the information extracted. 
The specific prompt we used in our experiments is in Appendix~\ref{app:prompt}

\section{Results}
Results on the performance of the models are reported in percentage, where in each experiment we evaluate the performance of select models on all instances of the \Turtle{} with test@1 method \cite{cobbe2021training} where the model generates only one piece of code for each instance. We assign a binary value to the success/failure of each task. We then run each experiment on a model five different times and we report the average percentage of accumulative success.

\begin{table*}[tbp]
  \centering
  \begin{tabular}{lccc|c|cc|c}
    \toprule
Task Type & \multicolumn{3}{c|}{\scratch \colorbox{colCg}{CG}} & \tweak \colorbox{colCg}{CG} &  \multicolumn{2}{c}{\tweak \colorbox{colCg}{CE}} & Runnable \\
Modalities & \colorbox{colIn}{T} & \colorbox{colIn}{I} & \colorbox{colIn}{I + T} & \colorbox{colIn}{I + T} & \colorbox{colIn}{I + T} & \colorbox{colIn}{I + I} \\

    \midrule
    GPT-4o/basic & 37.04 & 16.03 & \textbf{37.98} & 17.69 & 18.12 & 12.06 & 99.21\\
    GPT-4o/CoT & 38.12 & 19.23 & \textbf{40.18} & 20.00 & 19.61 & 13.84 & 99.85\\
    GPT-4o/4-S & NA & 21.49 & NA & NA & NA & NA  & 99.85 \\
    Gemini/basic & \textbf{25.09} & 7.71 & 22.22 & 3.85 & 12.00 & 3.00 & 99.13  \\
    Gemini/CoT & 18.51 & 9.20 & \textbf{20.52} & 7.70 & 23.08 & 11.54 & 99.17 \\
    Gemini/4-S & NA & 10.18 & NA & NA & NA & NA & 99.92 \\
    Llava/basic & \textbf{6.01} & 0.82 & 0.03 & 0.31 & 1.04 & NA & 69.13 \\
    Llava/CoT & \textbf{6.22} & 0.98 & 1.02 & 0.92  & 1.09 & NA & 72.34 \\
    \bottomrule
  \end{tabular}
  \caption{Performance on \Turtle{} (I = image, T = text, CG = code generation, CE = code edit, 4-S = 4-shot). Performance on \scratch tasks that include text (T) is calculated on a subset (21\%) of these tasks. Our result shows that while models' generated code is almost always runnable, they fail to generate desired shapes. }
  \label{tab:results-all}
  
\end{table*}

\subsection{Models perform poorly on \Turtle}

We initially examine the performance of the \gpt, \gemini and \llava{} models on the comprehensive \Turtle dataset. 
The findings, detailed in Table ~\ref{tab:results-all}, reveal a notably poor performance across the tasks in \Turtle, with a peak accuracy of 20\% achieved by \gpt in the \emph{code editing} tasks, facilitated by Chain of Thought (CoT) prompting. 
In the \emph{scratch} tasks, which represent the simplest problem type within the dataset, \gpt's success rate was just 19\%, underscoring the substantial challenges and complexities these tasks pose to the current models. 
A comparison between CoT and basic prompting within Table ~\ref{tab:results-all} illustrates that CoT prompting outperforms basic prompting on the same models, aligning with previous work that indicates CoT enhances models' reasoning abilities \citep{zhang2023multimodal}. 
However, despite employing CoT, the task remains far from being solved.
Examples of model output in different subsets of the task are provided in Figures \ref{fig:eval_examples} and \ref{fig:example2}. 

As previous work suggests that in-context learning by providing examples of the task at hand can significantly improve models' performance in domain adaptation \cite{brown2020language}, we evaluated the performance of \gpt and \gemini on \scratch \colorbox{colCg}{code generation} with 4-shot CoT prompting where we first provided four pairs of shape-code to the model. Yet, we did not find any major improvement in the performance of the models.

\subsection{\includegraphics[width=0.6cm]{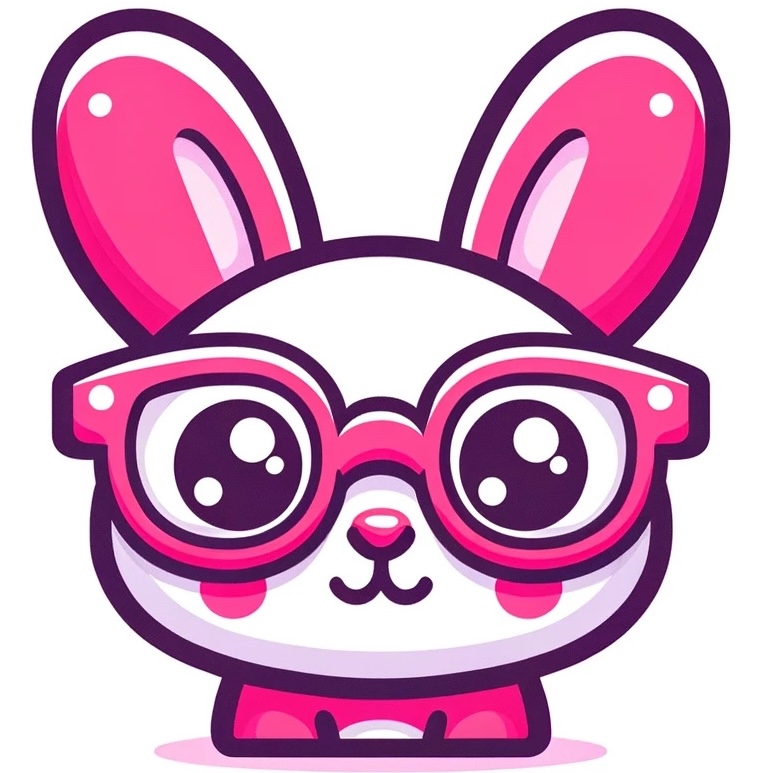}  Models fail to generalize}
\label{sec:rabbit}
Given that these models have been extensively trained on vast datasets sourced from the internet, there's an underlying uncertainty regarding the source of their performance—albeit poor—on the \Turtle tasks. 
Specifically, it remains unclear whether this performance is the result of the models' ability to memorize aspects of our tasks, rather than genuinely understanding and solving them based on their programming and reasoning capabilities.
To address this issue, our next step is to evaluate the true generalization ability of these models.
By doing so, we aim to distinguish between superficial learning, potentially influenced by memorization, and genuine comprehension and problem-solving skills. 
To measure the generalizability of the model's performance, we define an arbitrary set of commands based on the turtle module in Python.
In other words, we developed a class called \rabbit{} that inherits the \texttt{Turtle} class from the turtle module. 
Although the functions of the \rabbit{} class are functionally identical to those in the original turtle module, they are nominally distinct.
This differentiation allows us to evaluate the models' ability to apply their knowledge to unfamiliar yet equivalent command sets.
The definition of the \rabbit{} class in Python is provided in Appendix~\ref{app:rabbit}.
We perform a zero-shot CoT prompting to elicit the code using the new set of commands.
In the context window, we provide a verbal definition of each function in the \rabbit{} class.
The results of comparing the models' performance using the \rabbit{} class versus the standard Python Turtle module are presented in Table~\ref{tab:results-generalization}. 
We observe that, although both models were capable of generating executable pieces of code with the new class, there is a huge decline in their performance relative to their performance with the conventional Python Turtle module. 
This finding suggests that the visual reasoning in these models is not robust to syntax changes, and it is likely that they rely on training memorization rather than pure reasoning.

\begin{table}[tbp]
  \centering
  \begin{tabular}{lcccc}
    \toprule
     & \makecell{ \includegraphics[width=0.35cm]{figs/turtle.png} \\ \gpt} & \makecell{ \includegraphics[width=0.35cm]{figs/rabbit.jpeg} \\ \gpt}  & \makecell{\includegraphics[width=0.35cm]{figs/turtle.png} \\ Gemini} & \makecell{\includegraphics[width=0.35cm]{figs/rabbit.jpeg} \\ Gemini}\\
    \midrule
    \multicolumn{5}{l}{\scratch \colorbox{colCg}{Code Generation}} \\
    \colorbox{colIn}{I} & 19.23 & 6.00 & 9.20 & 3.00\\
    \midrule
    \multicolumn{5}{l}{\tweak \colorbox{colCg}{Code Generation}} \\
    \colorbox{colIn}{I + T}  & 20.00 & 5.04 &  7.70 & 2.37 \\
    \bottomrule
  \end{tabular}
  \caption{Performance of \gpt{} and \gemini on generalization tasks using CoT prompting, in these tasks. The performance in \rabbit drastically drops, showing poor generalization abilities in both models.}
  \label{tab:results-generalization}
\end{table}

\subsection{Assessing Model Proficiency Across Programming Languages}
\label{sec:beyond}
The initial suspicion might be that the models struggle with tasks in turtle geometry due to a lack of exposure to specific programming syntax during pretraining: Is the poor performance because of unfamiliarity with Python Turtle?

To answer the question, we run an ablation study on \gpt as our best-performing model in the main tasks.
We allow it to generate code using any library, language, or similar tools it deems appropriate, such as Matplotlib, TikZ, etc., without restricting it to the Python Turtle library. 
The prompt for this subset of tasks is presented in Appendix~\ref{app:arbitrary}.
We manually evaluate the \gpt output for this task.
Despite this freedom, we observe no significant improvement in performance. 
The model chooses Matplotlib for 50\% of the tasks and offers pseudocode for 2\%, with the remainder reverting to Python Turtle, even though we do not specify Python Turtle in the prompts. 
Notably, it avoids using TikZ, despite its mention in the prompt and proven capabilities in prior work to produce TikZ code \citep{bubeck2023sparks,belouadi2023automatikz}. 
We further isolate this observation by obligating the model to produce code using Matplotlib (refraining from generating pseudocode and Python Turtle) and in this experiment as well, we do not see a major improvement in the model's results.
This outcome underscores a deeper issue than syntax familiarity: the models' fundamental challenge is accurately interpreting visual input and applying this understanding to generate corresponding programming code.


\begin{table}[tbp]
  \centering
  \begin{tabular}{lccc}
    \toprule
    Output  & \makecell{Python\\Turtle} & \makecell{Any} & \makecell{Matplotlib} \\
    \midrule
    \multicolumn{4}{l}{\scratch \colorbox{colCg}{Code Generation}} \\
    \colorbox{colIn}{I}  & 19.23 & 21.6 & 22.15 \\
    \midrule
    \multicolumn{4}{l}{\tweak \colorbox{colCg}{Code Generation}} \\
    \colorbox{colIn}{I + T} & 12.3 & 15.11 & 15.23\\
    \bottomrule
  \end{tabular}
  \caption{
 Ablation study of CoT prompting of \gpt on code generation tasks where the model is given the freedom to choose the programming language. Producing code in Matplotlib does not yield qualitatively better performance. 
  }
  \label{tab:results}
\end{table}

\subsection{Limited Visual Understanding in LMMs: Insights from Textual vs. Visual Tweak Tasks}
To distinctly assess the models' proficiency in interpreting textual versus visual information, we conducted an evaluation focusing on their ability to reason about the relationship between provided code and corresponding images in \tweak \colorbox{colCg}{code edit} tasks.
In this setup, models are given a base shape along with the Python Turtle code that generated it.
Subsequently, they are prompted to adjust the given code to modify the shape according to specified instructions.
These instructions are delivered in two forms: 1) (\colorbox{colIn}{I + T}) as natural language descriptions, e.g., 'connect the midpoints of each side to the midpoints of its adjacent sides,' and 2) (\colorbox{colIn}{I + I}) as images explicitly showcasing the desired outcome.
Examples of these subsets are in Figure \ref{fig:intro}.
Our comparison of model performance across these two modalities reveals a huge decline in accuracy when instructions were provided visually rather than textually (Table~\ref{tab:results-all}). 
This outcome suggests a disparity in the models' ability to process visual versus textual instructions, revealing that their reasoning abilities may not align closely with human-like understanding. 
The assumption that directly viewing the desired outcome simplifies the task contrasts sharply with our findings, highlighting a reliance on textual interpretation for reasoning and a notable limitation in pure visual reasoning capabilities within these models \cite{roberts2024smart}.

\subsection{Vision component contributes poorly}

\label{sec:poor_visual}
One of the questions regarding LMMs' abilities in visual abstraction and understanding tasks is the extent the incorporation of the visual component has enhanced their abilities in reasoning \citep{mitchell2023comparing}. 

In resonance with what \citet{mitchell2023comparing} found, here we also found that the vision component contributes poorly to fostering the models' visual reasoning abilities, at least in the domain of \Turtle. Specifically, we annotated 27 (21\%) \scratch \colorbox{colCg}{code generation} tasks and provided clear descriptions for each in plain text (Textual descriptions were validated by two human annotators, one in the research team and the other recruited as a volunteer). The remaining shapes were too complex to describe without ambiguity in plain text.
Then, we compared the three modes of presenting the task, image only, text only, and the blend of an image and its textual description (\colorbox{colIn}{I}, \colorbox{colIn}{T}, and \colorbox{colIn}{I + T}, respectively in Table~\ref{tab:results-all}). Interestingly, for both \gpt and \gemini, the model performed worse when the task was presented only in the image, compared to the other modes. 
This phenomenon is counterintuitive as for humans, perceiving the images should be easier than first reading a description, imagining it, and then writing a code for it. 
Additionally, as presented in Table~\ref{tab:results-all} the blend of image and text only slightly improved \gpt's performance (from 38\% to 40\%). 
These two findings show that there is still much room for improvement especially in the visual components of LMMs.

\section{Related Work}
\subsection{Large Multi-modal Models} 
Recent advancements in foundational multimodal models have marked a significant stride towards developing generalist AI systems capable of understanding and integrating information across different modalities to solve tasks without the need for task-specific fine-tuning.
Among these models are closed source models such as \gemini \citep{geminiteam2023gemini}, \gpt \citep{openai2024gpt4}, and
open source models as LLaVA-1.5 \citep{liu2023improved}, Mini-GPT4 \citep{zhu2023minigpt4},  InstructBLIP \citep{dai2023instructblip} and CogVLM \citep{wang2024cogvlm}.
The versatility and multimodal understanding exhibited by these foundational multimodal models have positioned them as prime candidates for applications such as AI software engineers or programming tutors for children.
Our work evaluates the efficacy of these popular models on image/text-to-code tasks, measuring their potential in vision/programming context.

\subsection{Probabilistic Program Induction}
Recent work in Bayesian cognitive science has modeled various aspects of cognition and learning as probabilistic program induction \citep{lake2015human,lake2020people,rule2020child,ellis2023dreamcoder,wong2021leveraging,grand2023lilo}. This has involved both modeling human cognition as program induction as well as designing machine learning algorithms that can generate programs for various tasks, including the kind of turtle geometry task we study here. 
\citet{ellis2023dreamcoder} developed the DreamCoder algorithm which can learn to induce programs by using self-supervision to incrementally build up a library of programs and train a neural network to search to find the best program for a given task. They created a dataset of 160 turtle programming tasks. In contrast to our approach, where we assess the performance of out-of-the-box LMMs, DreamCoder is trained on a training set of images (i.e., half of the dataset). However, it is interesting that the algorithm is trained in an unsupervised fashion; that is, DreamCoder never receives the code used to generate the images and learns that from experience. \citet{wong2021leveraging} extended this work by developing an algorithm (LAPS) that can induce programs given both the task and linguistic annotations for the task. They used a dataset of 311 turtle graphics with greater complexity than the original DreamCoder dataset. While their dataset includes linguistic annotations, their dataset does not include tweak tasks like in \Turtle. Additionally, their tasks often include arbitrary aspects (for example, a gap with unspecified distance between two shapes) that makes evaluation hard; in our tasks, the positional relationships between shapes should be easy to infer exactly and hence we can evaluate models by comparing exactly with ground truth shapes. Moreover, neither of these datasets have been framed as a benchmark for visual program induction and have not been considered for evaluating LMMs. Perhaps the approach closest to our work is by \citet{grand2023lilo}, who combined LLMs with a symbolic program induction algorithm and evaluated the performance of their model (LILO) on the turtle geometry task using the aforementioned dataset. Averaged over several runs, the performance of the best versions of these approaches on the turtle geometry task is as follows: 43\% for DreamCoder, 82\% for LAPS, 49\% for LILO, and 32\% for a LLM solver. These results seem to suggest that probabilistic programming approaches (such as LAPS) can greatly outperform LMMs on visual programming tasks. We note that the performance of the LLM solver (32\%) is comparable to the performance of \gpt on our text-only input (37\%; see Table~\ref{tab:results-all}). Future work could assess the performance of probabilistic program induction methods like LAPS on \Turtle.

\subsection{Multimodal Reasoning}
The existing literature features a range of studies that evaluate these models using naturalistic images \citep{jiang2022bongard,johnson2017clevr,antol2015vqa}, yet humans naturally are able to reason over abstract shapes \citep{chollet2019measure,zhang2019raven,spelke2007core} and also many use cases of LMMs involve understanding abstract shapes and sketches \citep{forbus2011cogsketch,nie2020bongard}. Moreover, unlike naturalistic images \citep{marjieh2022words,sucholutsky2024alignment}, the relationship between language and abstract shapes is highly intertwined as minimal alterations in language can lead to different visual perceptions in humans \citep{dldillon, lin2023we}. Overall, recent surveys on deep learning for mathematical reasoning \cite{lu2022survey,sun2023survey} have pointed out that most of the available datasets sand benchmarks on multimodal reasoning often rely on visual question-answering frameworks. However, these methods fall short because they are usually trained on datasets composed of natural images, rather than on datasets tailored to the integration of vision and language for mathematical tasks. 

The Multimodal Algorithmic Reasoning (MAR) task tests multi-modal models on fundamental skills understandable by children, focusing on interpreting visual and linguistic information to answer questions. Perhaps the most relevant work to ours is the paper by \cite{cherian2023deep} in which they introduced a dataset with 101 multiple-choice questions inspired by the Math Kangaroo contest for 6 to 8-year-olds, involving images and texts that the model must analyze together. The task has been shown to be challenging for multimodal deep neural networks, and the following trials to solve the problem have gained less than 25\% accuracy on the test set \cite{wu2023solution}.
\Turtle includes abstract geometric shapes, and the task only relies on knowledge and reasoning over a set of simple functions in the Python Turtle library. Our open-ended benchmark and its flexibility over different modalities make evaluating different aspects of multimodal reasoning in LMMs more reliable. 

\section{Discussion on Educational Implications}

While LLMs and LMMs have sparked interest among education researchers in AI tools such as tutors \cite{MeetKhan43:online} for students, our work cautions against using these models without thorough evaluation. Although students can learn turtle programming on various platforms without AI help (e.g., \citet{Codewith82:online,SelfPace49:online}), the effectiveness of these models as tutors or copilots is uncertain due to current limitations. Our work suggests that educational researchers can engage in systematically benchmarking LMMs to ensure their reliability before integrating them into student learning processes.

\section{Conclusions}
This study introduces \Turtle, the first of its kind in benchmarks that focus on converting visual inputs to code outputs.
The evaluation results from \Turtle reveal a significant disparity between how humans tackle turtle programming and how SOTA AI models perform in understanding simple geometric shapes, reasoning about these shapes, and converting such understandings into executable code \cite{rismanchian2024turtle}. 
This gap underscores the challenges that lie ahead in the quest to enhance AI's comprehension and problem-solving abilities to match human levels.
We believe that \Turtle serves as a crucial tool in the evaluation of models, offering a clear benchmark testing the limits of LMMs.

\section{Limitations}
One of the limitations of our work is that we did not experiment with fine-tuning techniques to better understand multimodal reasoning abilities in these models and how they could be improved. However, we argue that our experiments demonstrate that poor performance in the models is perhaps not due to their unfamiliarity with the syntax of Turtle, but rather is more related to vision components and their reasoning abilities. We plan to experiment with fine-tuning techniques in future work with smaller models such as \llava{} with newer architectures for vision towers \cite{li2024tackling} to examine the effectiveness of these techniques. 

Finally, as \Turtle has a limited number of instances, future work can augment existing instances to produce a dataset plausible for training purposes.

\section*{Acknowledgements}
We are grateful to Arghavan Rezvani for her valuable assistance and insightful contributions.

\bibliography{custom}

\appendix
\section{Appendix}
\subsection{Reasons of Failure}
We manually investigated \gpt's failures in solving Scratch tasks in a single run to find the major causes of failure. 
We find four major causes: 1) Shape identification error: where the model fails to completely capture existent shapes in the input image, for instance, if it confuses a semicircle with a circle or assigns non-existent shape attributes to the input image. 2) Counting error: where the model fails to count adequately, (e.g., three triangles counted as four), 3) Orientation error: where the model fails to correctly find the relationships between different components of a shape (e.g., semicircle on top of a square vs. at its bottom), and 4) Implementation error: where the model's generated code does not follow the pre-planned pseudocode.

\begin{table*}
\centering
\begin{tabular}{p{3.5cm}p{7.5cm}c}

\textbf{Cause} & \textbf{Description} & \textbf{Percentage} \\
\hline
Shape identification error & The model fails to completely capture existent shapes in the input image, confusing or misattributing shapes  & 25\% \\
\hline
Counting error & The model inadequately counts the elements. & 35\% \\
\hline
Orientation error & The model fails to correctly determine the spatial relationships between different components of a shape  & 21\% \\
\hline
Implementation error & The model's generated code does not adhere to the pre-planned pseudocode, resulting in incorrect implementation. & 45\% \\

\end{tabular}
\caption{Major Causes of \gpt's Failures in Scratch Tasks; note that the failures are not mutually exclusive, as a model can perform a combination of errors in each task}
\label{tab:gpt_failures}

\end{table*}

We manually investigated \gpt's failure output in the scratch code generation task and the results are provided in Table~\ref{tab:gpt_failures}, where the failures are not mutually exclusive as a model can perform a combination of errors in each task. Furthermore, while the first three errors are according to the vision component in these models, we see that 64\% of the failures are according to these causes, and in 36\% of failure cases, there are no apparent vision errors.

\subsection{Prompting}
\label{app:prompt}
\subsubsection{Basic Prompt}
\begin{figure}[h]
    \centering
        \begin{minted}[fontsize=\footnotesize, frame=single,linenos=false,breaklines,breaksymbol=,escapeinside=||,bgcolor=white]{text}
In each task, the user provides an image of an abstract geometric shape or pattern and an instruction, you need to generate a code in Python Turtle that follows the user's request. 
\end{minted}
\caption{basic prompt used in our experiments}
\end{figure}

\subsubsection{v-CoT Prompt}
\begin{figure*}[h]
    \centering
        \begin{minted}[fontsize=\footnotesize, frame=single,linenos=false,breaklines,breaksymbol=,escapeinside=||,bgcolor=white]{text}
You are Turtle Geometrician, you are an expert in reasoning about images and generating code in Python Turtle using images You need to follow the steps below before generating the answer:
(1) Describe the relevant information from the image needed to answer the question. List all relevant artifacts from the image.
(2) Use the information described in (1) to reason about the problem by working step by step to arrive at the final piece of code.
(3) Generate the final code. NEVER use "pensize" function in your code.
\end{minted}
\caption{v-CoT prompt used in our experiments}
\end{figure*}
\subsubsection{A Complete Example}
\begin{figure*}[!ht]
    \centering
    \begin{minipage}{.1\textwidth}
        \centering
        \includegraphics[width=\linewidth]{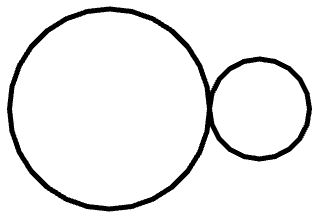}
    \end{minipage}%
    \begin{minipage}{.9\textwidth}
        \centering
        \begin{minted}[fontsize=\footnotesize, frame=single,linenos=false,breaklines,breaksymbol=,escapeinside=||,bgcolor=white]{text}
System: You are Turtle Geometrician, you are an expert in reasoning about images and generating code in Python Turtle using images. You need to follow the steps below before generating the answer:
(1) Describe the relevant information from the image needed to answer the question. List all relevant artifacts from the image.
(2) Use the information described in (1) to reason about the problem by working step by step to arrive at the final piece of code.
(3) Generate the final code. NEVER use "pensize" function in your code.

Text: Provide a code in Python turtle that in the given shape inserts a circle of an equal size to the smaller circle on the left of the bigger circle to make a vertically symmetrical shape.

Complete the code: 

import turtle
from math import *
t = turtle.Turtle()
large_circle_radius=100
small_circle_radius=50
...
        \end{minted}
    \end{minipage}
    \caption{An example of a complete prompt for a tweak code generation task with using v-CoT prompting.}

\label{fig:comp_example}
\end{figure*}

Figure~\ref{fig:comp_example} we provide an instance of a complete prompt we used for a \colorbox{colSc}{tweak} \colorbox{colCg}{\emph{code generation}} task with CoT prompting.

\subsubsection{Arbitrary Output}
\label{app:arbitrary}
Figure~\ref{fig:arbitrary} provides the CoT prompt we used for the model to provide a code in any arbitrary language or library that creates the desired shape.

\begin{figure*}[h]
    \centering
        \begin{minted}[fontsize=\footnotesize, frame=single,linenos=false,breaklines,breaksymbol=,escapeinside=||,bgcolor=white]{text}
You are an expert in reasoning about images and generating code in any language you prefer. You need to follow the steps below before generating the code that answers the user's request:
(1) Describe the relevant information from the image needed to answer the question. List all relevant information from the image.
(2) Use the information described in (1) to reason about the problem by working step by step to arrive at the final piece of code.
(3) Generate the final code. Your code can be in any visual language or library, such as Matplotlib, TikZ, etc.
\end{minted}
\caption{The system prompt we used for the results discussed in Section~\ref{sec:beyond}}
\label{fig:arbitrary}
\end{figure*}

\subsection{\rabbit{}}
\subsubsection{Prompt used}
\begin{figure*}[h]
    \centering
        \begin{minted}[fontsize=\footnotesize, frame=single,linenos=false,breaklines,breaksymbol=,escapeinside=||,bgcolor=white]{text}
Suppose that I have a library named Rabbit in Python. Rabbit library has an object constructor named Rabbit which is an object that moves on the screen and draws lines. It only has these functions:
aa(length): goes front or back (if the length is negative) and draws a line with the length of pixels.
bb(degree): The rabbit turns its head right or left (if degree is negative).
cc(radius, degree): creates an arc with the given radius for the given degree. If degree=360 it creates a circle. The center of the circle is in the left of the rabbit.
pp(vanish): if vanish=True vanishes Rabbit object so if it moves does not draw anything, and if vanish=False, it appears the Rabbit object so if it moves draws on the screen. 
you call the functions on an object of Rabbit, such as r.aa(length) where r is an object of Rabbit. When r is created, it faces north (up) on the screen and it does not vanish, so it is in drawing mode. 

You are Rabbit Geometrician, you are an expert in reasoning about images and generating code in Python Rabbit using images. You need to follow the steps below before generating the answer:
(1) Describe the relevant information from the image needed to answer the question. List all relevant artifacts from the image.
(2) Use the information described in (1) to reason about the problem by working step by step to arrive at the final piece of code.
(3) Generate the final code. Only use commands in the Rabbit class.
\end{minted}
\caption{v-CoT prompt used for generalization experiments discussed in Section~\ref{sec:rabbit}}
\label{fig:rabbit_prompt}
\end{figure*}

The prompt we used for this experiment is provided in Figure~\ref{fig:rabbit_prompt}.

\subsubsection{Definition of the class}
The rabbit class is an arbitrary class that we defined based on \textit{Turtle} class in the Python Turtle Module. This minimal set of functions includes all functions that a programmer or a model needs to create all of the tasks in \Turtle. We defined this new set of functions to measure how \gpt is able to generalize its abilities in generating code in Python Turtle to a similar but minimally different set of functions. 

\label{app:rabbit}
\begin{lstlisting}[language=Python]
import turtle

class Rabbit(turtle.Turtle):
    def __init__(self):
        super().__init__()
        self.setheading(90)
        self.pensize(5)
        self.hideturtle()

    def aa(self, length):
        self.forward(length)

    def bb(self, degree):
        self.right(degree)

    def cc(self, radius, degree):
        self.circle(radius, degree)

    def pp(self, vanish):
        if vanish:
            self.penup()
        else:
            self.pendown()    
\end{lstlisting}

\subsection{Types of Tweak Tasks}
\label{app:tweaks}
\begin{figure*}
    \centering
    \includegraphics[width=0.8\linewidth]{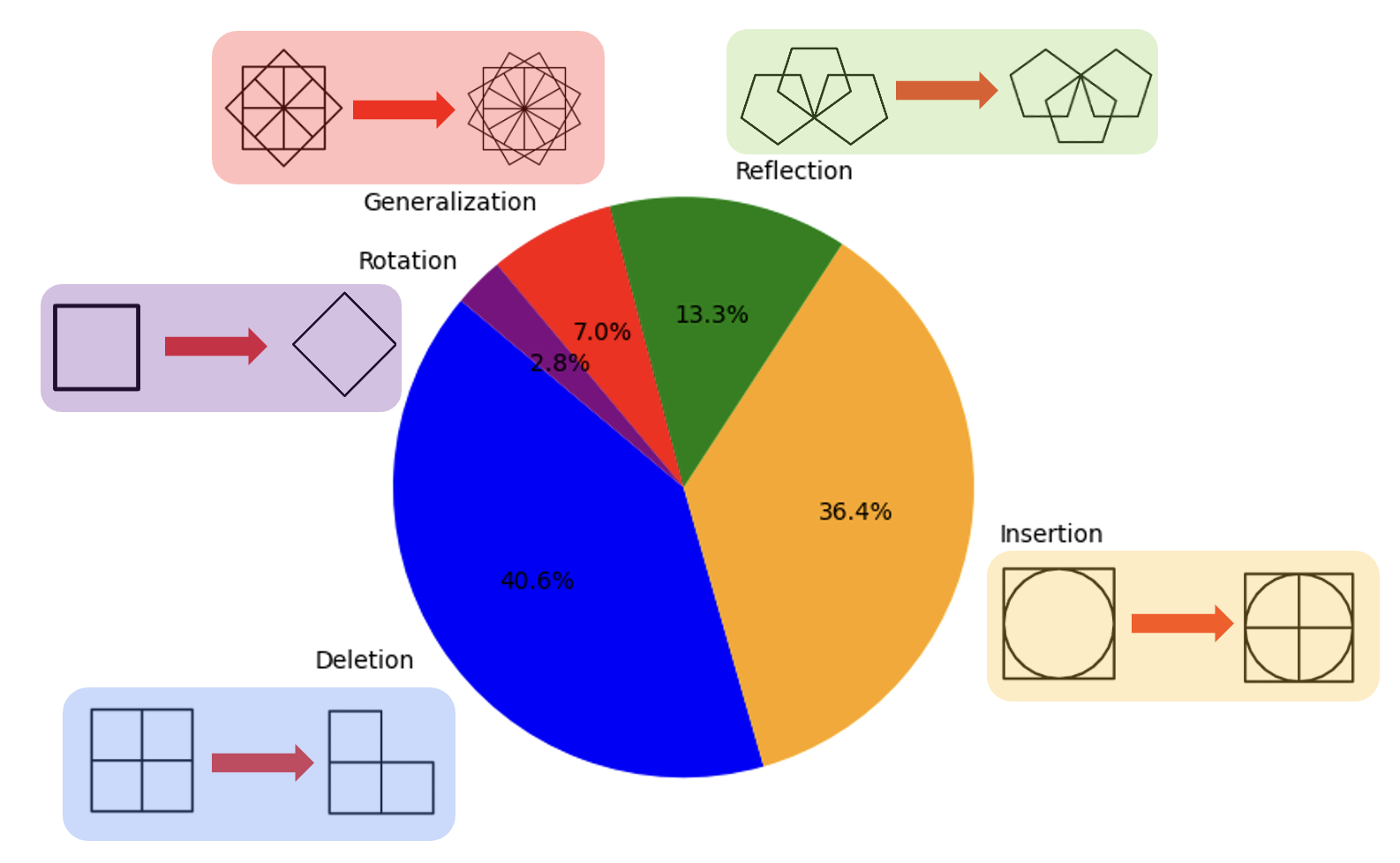}
    \caption{Types of tweaks and their share in \Turtle}
    \label{fig:types_of_tweaks}
\end{figure*}

\Turtle includes a total of $130$ tweak tasks.
We provide a categorization for the tweaks as follows:
There are five major types of tweaks in \Turtle{};
\begin{itemize}
\item{Deletion}: Removing a specified part of a shape
\item{Insertion}: Adding a specific shape to the pattern as directed
\item{Rotation}: Rotating the entire shape
\item{Reflection}: Reflecting the entire shape or parts of it across specified lines
\item{Generalization}: maintaining a pattern in the image constant while varying its parameters.
\end{itemize}
An illustration of instances of each type is provided in Figure~\ref{fig:types_of_tweaks}. 
These types are not mutually exclusive as 10\% of the tasks involve a combination of two types (e.g., removing one side of a square and inserting a semicircle instead).
To successfully complete deletion and insertion tweaks, a model needs to demonstrate a nuanced understanding of the details in the image and program the resulting shape accordingly. 
In contrast, rotation tasks can be relatively easy as most of them can be solved only using a simple function in Turtle that can rotate the starting heading of the turtle which results in complete rotation in the entire shape (i.e., \texttt{turtle.right(angle)}).

\subsection{Evaluating Image Complexity Using Contour Counts}
As our result suggests that the vision component is contributing poorly to the models' performance, to gain a better understanding of the visual obstacles for the models to solve the tasks, we defined a measure as a proxy for the complexity of shapes. 
For each provided image, we calculated the number of contours in each shape.
In OpenCV, a contour is a curve joining all the continuous points (along the boundary), having the same color or intensity. 
Contours are a useful tool for shape analysis and object detection and recognition.
The high number of contours in an image hints that there are many shapes being involved and interleaving with each other, which makes understanding and extracting underlying patterns challenging.

We calculated the number of contours in each shape by utilizing the corresponding function in OpenCV,
and defined three arbitrary levels of complexity in the images, where the images which include only one contour (e.g., the basic square in Figure~\ref{fig:intro}) are at level 1 (simple), images including less than 6 contours and more than 1 are at level 2 (medium) (e.g, the base shape of insertion example in Figure~\ref{fig:types_of_tweaks}) and the images in which there are more than 6 contours (e.g., the base shape in generalization example in Figure~\ref{fig:types_of_tweaks}) are at level 3 of the complexity (Complex).  
In Turtle, the proportions of complexity levels 1, 2, and 3 are 25\%, 40\%, and 35\%, respectively.

We investigate how models perform over tweak tasks. There are 9 different ways that a pair of input and output image can combine. As shown in Table~\ref{table:complexity}, the majority of tweak tasks (74) have same levels of complexity for the input and output image. \\
To examine how complexity of input and output shapes impact the results, we categorize tweak tasks in the 9 different categories and count the number of tasks that are ever solved by \gpt under any prompting method in code generation and code edit tasks during 6 different runs. As shown in Table~\ref{table:complexity}, the more complex the input shape is, the more challenging solving the task is.

\begin{table*}[h]
\centering
\begin{tabular}{ccccc}
\multicolumn{2}{c}{\multirow{2}{*}{}} & \multicolumn{3}{c}{Output Complexity} \\ \cline{3-5} 
\multicolumn{2}{c}{}                  & Simple & Medium & Complex \\ 
\multirow{3}{*}{Input Complexity} & Simple   & \textcolor{green!60!black}{35\% (7/20)} & \textcolor{green!70!black}{30\% (3/10)} & \textcolor{yellow!60!green}{25\% (1/4)} \\  
                                  & Medium   & \textcolor{green!60!black}{40\% (2/5)}  & \textcolor{yellow!65!red}{18\% (6/33)} & \textcolor{yellow!10!red}{7\% (1/13)} \\ 
                                  & Complex  & \textcolor{yellow!70!red}{20\% (1/5)}  & \textcolor{yellow!30!red}{11\% (2/19)} & \textcolor{yellow!60!red}{19\% (4/21)} \\ \hline
\end{tabular}
\caption{The number of tweak tasks under each category and the percentage of those tasks ever solved by \gpt in different settings.}
\label{table:complexity}
\end{table*}

\begin{figure*}
    \centering
    \includegraphics[width=\linewidth]{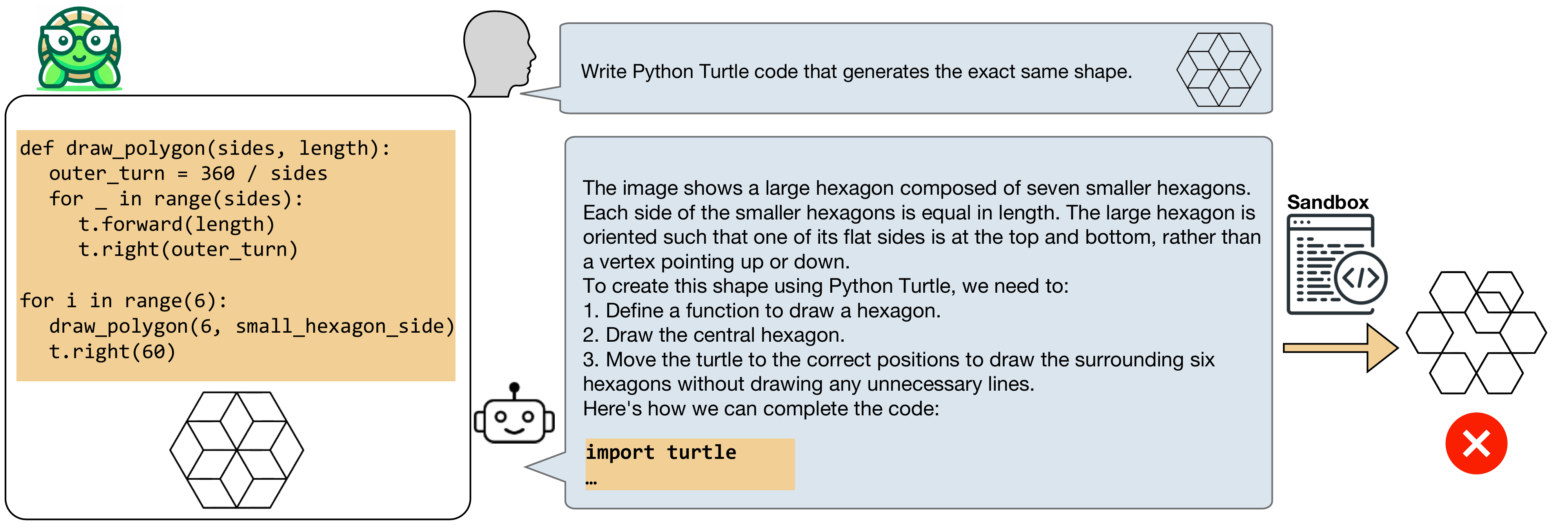}
    \caption{An illustration of our evaluation pipeline}
    \label{fig:eval_pipeline}
\end{figure*}

\begin{figure*}
    \centering
    \includegraphics[width=\linewidth]{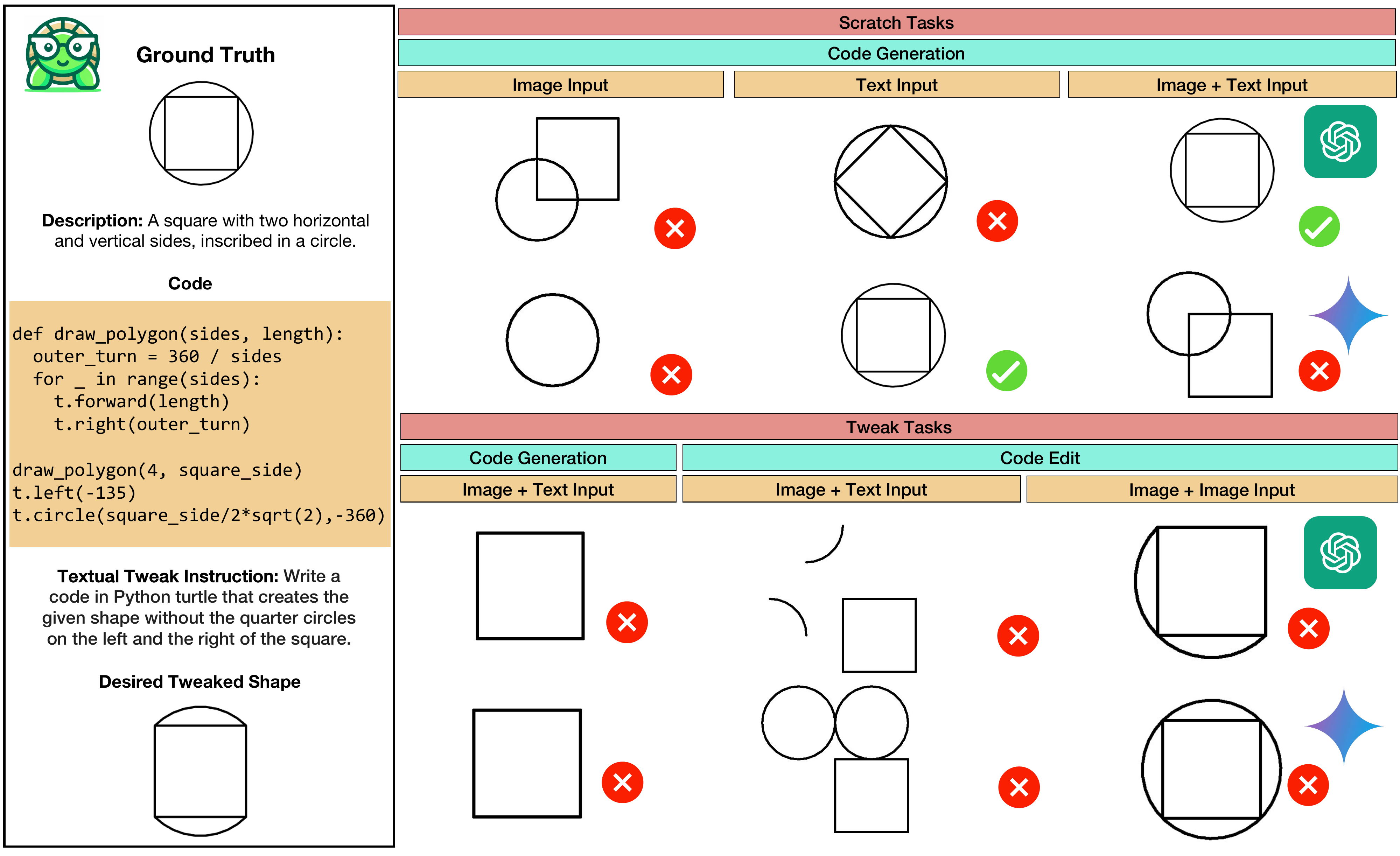}
    \includegraphics[width=\linewidth]{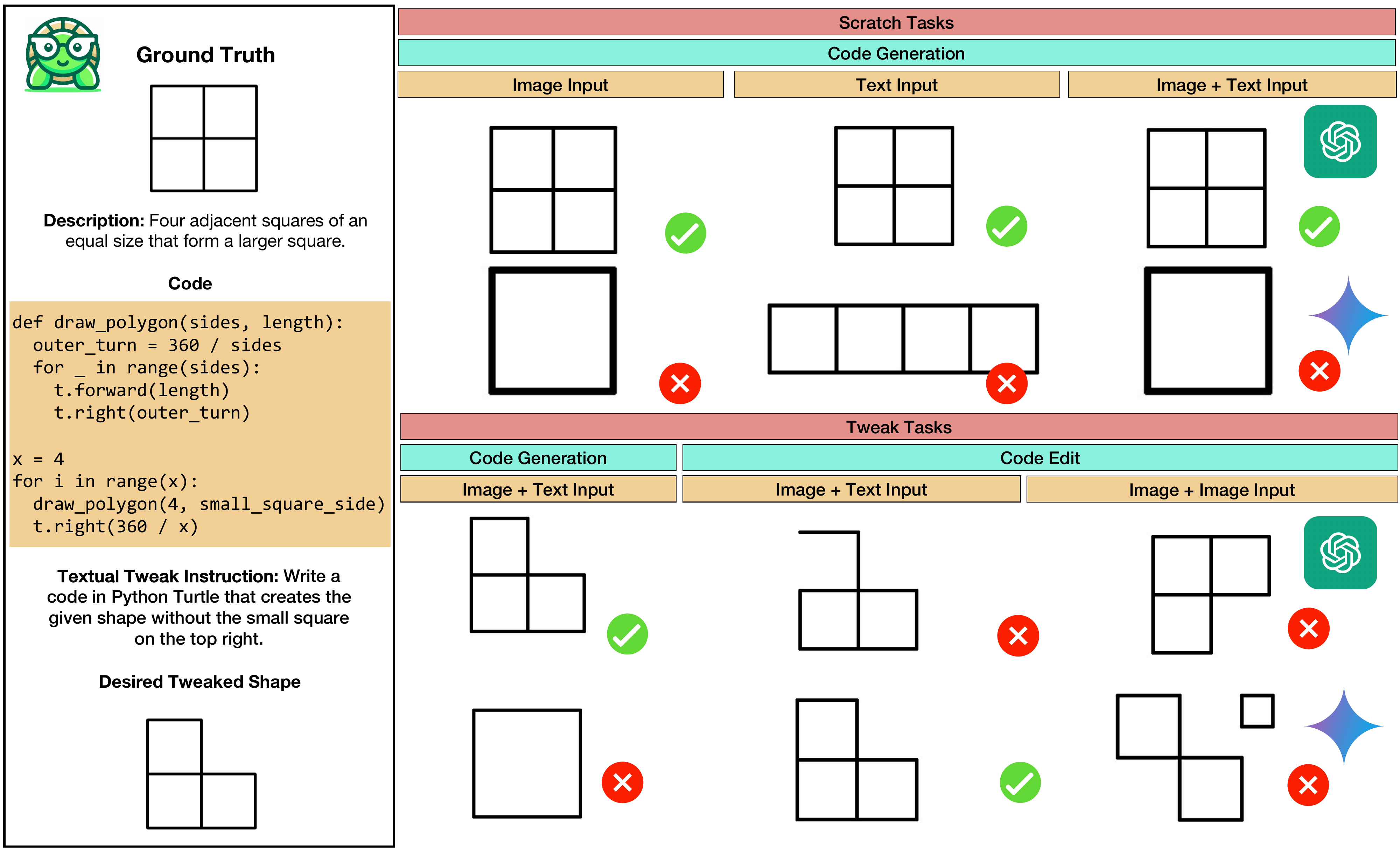}
    \caption{Two examples of tasks in \Turtle across different modalities}
    \label{fig:example2}
    
\end{figure*}


\end{document}